\documentclass[10pt,twocolumn,letterpaper]{article}

\usepackage[pagenumbers]{cvpr} 

\usepackage[dvipsnames]{xcolor}

\definecolor{cvprblue}{rgb}{0.21,0.49,0.74}
\newcommand{\mypar}[1]{\noindent\textbf{#1}}
\usepackage[pagebackref,breaklinks,colorlinks,citecolor=cvprblue]{hyperref}
\usepackage[capitalize]{cleveref}
\usepackage{algorithm, algpseudocode}
\usepackage{times}
\usepackage{latexsym}
\usepackage{subfiles}
\usepackage{graphicx}
\usepackage{booktabs}
\usepackage{diagbox}
\usepackage{slashbox}
\usepackage{float}
\usepackage{textcomp}
\usepackage{enumitem}
\usepackage{xcolor}
\usepackage{colortbl}
\usepackage{multirow}
\usepackage{subcaption}

\crefname{section}{Sec.}{Secs.}
\Crefname{section}{Section}{Sections}
\Crefname{table}{Table}{Tables}
\crefname{table}{Tab.}{Tabs.}

\def\paperID{2713} 
\def\confName{CVPR}
\def\confYear{2024}

\usepackage[dvipsnames]{xcolor}

\newcommand{\fg}[1]{\textcolor{red}{\textbf{FG: #1}}}

\newcommand{\secref}[1]{Section \ref{#1}}

\usepackage{acronym}
\acrodef{cv}[CV]{Computer Vision}
\acrodef{nlp}[NLP]{Natural Language Processing}
\acrodef{vl}[VL]{Vision-Language}
\acrodef{vln}[VLN]{Vision-Language Navigation}
\acrodef{vlm}[VLM]{Vision-Language Models}
\acrodef{ml}[ML]{Machine Learning}
\acrodef{t2i}[T2I]{Text-to-Image}
\acrodef{mm}[MM]{Multi-Modality}
\acrodef{nerf}[NeRF]{Neural Radiance Fields}
\acrodef{llm}[LLM]{Large Language Model}
\acrodef{genai}[GenAI]{Generative AI}
\acrodef{ddm}[DDM]{Denoising Diffusion Model}
\acrodef{sam}[SAM]{Segment Anything}
\acrodef{mdp}[MDP]{Markov Decision Process}
\acrodef{pomdp}[POMDP]{Partially Observable Markov Decision Process}
\acrodef{rl}[RL]{Reinforcement Learning}
\acrodef{il}[IL]{Imitation Learning}
\acrodef{bc}[BC]{Behavior Cloning}

\usepackage{xspace}
\makeatletter
\DeclareRobustCommand\onedot{\futurelet\@let@token\@onedot}
\def\@onedot{\ifx\@let@token.\else.\null\fi\xspace}
\def\eg{\emph{e.g}\onedot} 
\def\ie{\emph{i.e}\onedot} 
 
\def\etc{\emph{etc}\onedot}

\algnewcommand{\Inputs}[1]{%
  \State \textbf{Inputs:}
  \Statex \hspace*{\algorithmicindent}\parbox[t]{.8\linewidth}{\raggedright #1}
}
\algnewcommand{\Initialize}[1]{%
  \State \textbf{Initialize:}
  \Statex \hspace*{\algorithmicindent}\parbox[t]{.8\linewidth}{\raggedright #1}
}

\title{Planning as In-Painting: A Diffusion-Based Embodied Task Planning Framework for Environments under Uncertainty}

\author{
Cheng-Fu Yang$^{\spadesuit}$ \:\: 
Haoyang Xu$^{\spadesuit}$ \:\: 
Te-Lin Wu$^{\spadesuit}$ \:\: 
Xiaofeng Gao$^{\clubsuit}$ \:\: 
Kai-Wei Chang$^{\spadesuit}$ \:\: 
Feng Gao$^{\clubsuit}$ \\
$^\spadesuit$ University of California, Los Angeles \qquad 
$^\clubsuit$ Amazon\\
{\tt\small \{cfyang, ericxu19, telinwu, kwchang\}@cs.ucla.edu, \{gxiaofen, fenggo\}@amazon.com}\thanks{This work is not related to these authors' position at Amazon.}}

\begin{document}
\maketitle

\begin{abstract}
Task planning for embodied AI has been one of the most challenging problems where the community doesn't meet a consensus in term of formulation. In this paper, we aim to tackle this problem with a unified framework which consists of an end-to-end trainable method and a planning algorithm. Particularly, we propose a task-agnostic method named ``planning as in-painting''. In this method, we use a \acf{ddm} for plan generation conditioned on both language instructions and perceptual inputs under partially observable environments. Partial observation often leads to the model hallucinating the planning. Therefore, our diffusion-based method jointly model both state trajectory and the goal estimation in order to improve the reliability of the generated plan given the limited available information at each step. To better leverage newly discovered information along the plan execution for higher success rate, we propose an on-the-fly planning algorithm to collaborate with the diffusion-based planner. The proposed framework achieves promising performances in various embodied AI tasks, including vision-language navigation, object manipulation, and task planning in a photo-realistic virtual environment. \footnote{The code is available at: \href{https://github.com/joeyy5588/planning-as-inpainting}{github.com/joeyy5588/planning-as-inpainting}.}

\end{abstract}

\section{Introduction}
\label{sec:r-intro}

In recent years, embodied AI has surged in prominence across the fields of \acf{cv}, \acf{vl}, \acf{ml} and Robotics~\cite{kolve2017ai2, anderson2018vision, shridhar2020alfred, szot2021habitat, pashevich2021episodic, brohan2022rt, yang2023lacma}.
While a robot can be programmed to function autonomously in the wild, being able to receive and follow instructions to accomplish user-intended tasks (\eg, navigating to targeted place, manipulating certain objects, \etc) could lead to a more widely applicable and helpful assistive agent.

The agent must be able to effectively utilize both perceived environmental sensory inputs (primarily visual) and comprehended instructions (in the form of language) in order to function appropriately within the environment.


\begin{figure}[t]
    \centering
    \includegraphics[width=\linewidth]{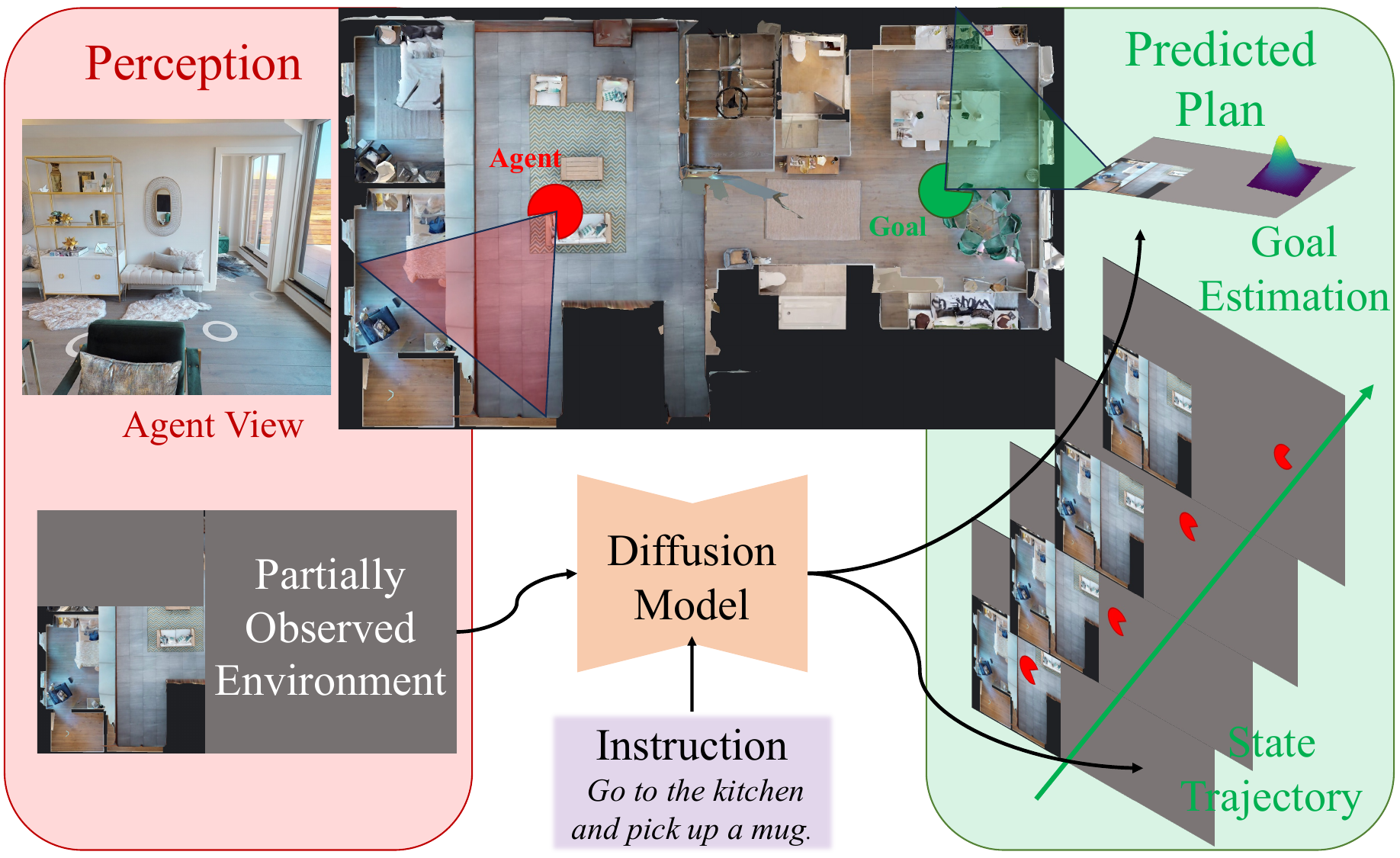}
    \caption{An illustration of our ``planning as in-painting'' formulation in a photo-realistic embodied environment.} 
    \label{fig:figure1}
    \vspace{-1.em}
\end{figure}

Practically, an instruction-guided agent can be approached by learning a conditional policy (\ie, conditioned on the instruction) where the induced plans (trajectories) either maximize the return from the environment (typically by achieving the instructed goal), or mimic expert behaviors when available.
In order to learn such a policy, conventional approaches leverage reinforcement learning (RL)~\cite{zhu2017target, janner2018representation, co2018guiding, sun2019program}, imitation learning~\cite{duan2017one, wang2017robust, das2018embodied, yang2023lacma, pashevich2021episodic}, as well as modularized frameworks combining the two aforementioned and other domain-specific modules~\cite{min2021film, inoue2022prompter, blukis2022persistent}.
On the other hand, contemporary research has made efforts to apply conditional generative models in order to tackle the problem of policy learning and decision making~\cite{janner2022planning, ajay2022conditional, chi2023diffusion}.
Specifically, these works adapt diffusion models~\cite{ho2020denoising} to learn to ``recover'' a set of reward-labeled trajectories through the generative process.
Compared to the conventional approaches, generative policy is demonstrated to suffer less from producing myopic actions, and enjoy more flexibility in incorporating constraints (modeled as conditions) that a learned policy should satisfy.
Despite these benefits, whether the generative policies can function properly under environment uncertainty, still remains an open question~\cite{ajay2022conditional}.
Considering the agent and the instruction it is supposed to follow in \autoref{fig:figure1}, the goal here, may not be fully observable to the agent.
In this scenario, a generative policy can risk hallucinating future decisions due to the partial observability~\cite{ajay2022conditional, ortega2021shaking} and hence fail to robustly follow the instructions.

In this work, we aim to address the aforementioned limitation of generative policy. In particular, we jointly model an \textit{intended goal estimation} as well as \textit{an actual plan} reaching such a goal with a language-conditioned \acf{ddm}.
Inspired by~\citet{ajay2022conditional}, our method directly models the trajectory (and hence the \textit{plan}) in the state-space. We can naturally model \textit{intended goal estimation} in the same space, \eg, as a belief goal state inducted by the instruction.
Hence, generating such a \textit{intended goal estimation} alleviates the uncertainty caused by the partially observable environment.
The plan, as a generated state-sequence, can then be ``in-painted'' between a present state and the belief goal state, along with other essential environment substances.

We thus propose a novel language-guided generative policy named ``planning as in-painting'', which uses a conditional diffusion model to train instruction-following agent under environment uncertainty.
In order to better handle environment uncertainty in a dynamic manner, we also introduce an \textit{on-the-fly} planning algorithm that reduces the dependence of environmental inductive biases as well as balances between policy exploration and exploitation more efficiently.

We evaluate our proposed method in three different simulated agent environments, including visual-language navigation, robotic arm object manipulation, and task planning in realistic egocentric environment. 
Our experimental results show the effectiveness of the proposed framework in all three evaluated environments, compared to previous methods including RL-based baselines, prior diffusion-based generative policy and modular methods. Furthermore, extensive ablation analysis demonstrates the advantages of both our ``planning as in-painting'' modeling and the \textit{on-the-fly} planning algorithm.

In summary, the contribution of our work are three-folds:
\begin{itemize}
    \item We propose a novel language-conditioned generative policy, dubbed ``planning as in-painting'', which is able to produce effective planning in partially observable environments.
    \item We introduce an on-the-fly planning algorithm, collaborating with the diffusion-based planner for better exploration and exploitation balance.
    \item We conduct a series of experiments as well as additional analysis to verify the effectiveness of the proposed method. 
\end{itemize}

\section{Related Work} \label{sec:related}

\mypar{Conventional Approaches in Embodied AI.}
In the realm of Embodied AI, several core tasks have attracted researchers' attention. These tasks typically include object manipulation~\cite{schreiber2010fast, fan2018surreal, james2020rlbench, jiang2022vima, ehsani2021manipulathor} and visual navigation~\cite{anderson2018vision, kipf2019compositional, shridhar2020alfred, krantz2020beyond, gao2023alexa, ku2020room}. 

For object manipulation, the diversity of tasks requires agents to acquire a versatile skill set tailored to specific task requirements. Thus, specialized solutions have been developed for instruction following~\cite{stepputtis2020language, jiang2022vima}, constraint satisfaction~\cite{brunke2022safe, schreiber2010fast, srinivasan2020learning, ajay2022conditional}, rearrangement~\cite{RoomR, gan2021threedworld, batra2020rearrangement}, tabletop manipulation~\cite{zhu2020robosuite, fan2021secant, jiang2022vima}, and goal-conditioned planning~\cite{srinivas2018universal}; each designed to equip agents with the capabilities needed to tackle the particular challenges posed by different object manipulation scenarios

Visual navigation, on the other hand, requires agents to navigate in complex environments using natural language instructions~\cite{ke2019tactical, chen2021history, jain2019stay, tan2019learning, zhu2020babywalk, li2019robust, zhu2021self, schumann2022analyzing}.
Techniques such as sequence-to-sequence models~\cite{pashevich2021episodic, suglia2021embodied, yang2023lacma} demonstrate strong navigation capability in the seen environment, yet their performance can falter in novel environments.
On the other hand, modular approaches~\cite{min2021film, inoue2022prompter, blukis2022persistent} learn to model the semantic mapping to mitigate the visual discrepancies across environments.
However, the disjoint nature between the visual module and the language planning module can hinder the agents from executing complex, multi-step instructions.
Our proposed framework aims at bridging the efficiency of planning in the feature space and the strong data distribution modeling capability of conditional generative models, to devise a more efficient planning framework that better utilizes a given instruction.

\vspace{.3em}

\mypar{Foundation Models in Embodied AI.}
The remarkable performance in both language (sequence) understanding and generation draws rapidly growing attention of Large Language Models (LLMs), even in the computer vision and robotics communities.
Increasingly more recent works have integrated LLM into a variety of solutions for addressing embodied tasks, with a particular focus on reasoning and planning.
Works such as~\cite{ahn2022can, huang2023grounded, yao2022react} have successfully anchored the high-level semantic knowledge of LLMs into executable low-level actions.
Meanwhile, numerous works~\cite{driess2023palm, brohan2023rt, shridhar2022cliport, shridhar2023perceiver, chen2023open, huang2023visual, shafiullah2022clip, fan2022minedojo} exploit Vision-and-Language Models (VLMs) to achieve either task or object generalization.
Lastly, there are also abundant literature~\cite{huang2022inner, huang2022language, liang2023code, li2022pre} that utilize LLMs for effective long-horizon planning. While these methods demonstrate proficiency in tasks within the well-explored environment, how these methods can cope with real-world partial observability, still remains an open research question.
To this end, our method proactively addresses this environmental uncertainty by incorporating the belief prediction.
This enables our model to infer the unobserved parts of the environment during test time and dynamically adjust its strategies, effectively handling the complexities of real-world environments.

\vspace{.3em}

\mypar{Generative Methods in Embodied AI.}
To enhance agent's learning, a series of studies have suggested modeling the environment's dynamics with a world model, as seen in the contributions of~\cite{hafner2019dream, ha2018world, hafner2020mastering, koh2021pathdreamer}. Recently, diffusion models have shown remarkable capabilities in modeling and generating high-dimensional data, as evidenced by~\cite{ho2020denoising, nichol2021improved, rombach2022high, nichol2021glide, zhang2023adding}.
Building on top of this, research works~\cite{chi2023diffusion, reuss2023goal, sridhar2023nomad, wang2022diffusion, pearce2023imitating} have leveraged diffusion models to deal with the inherent stochasticity in complex policy landscapes. Concurrently, certain studies~\cite{ajay2022conditional, dai2023learning, janner2022planning} have shifted focus to modeling state trajectories, a method that accounts for the typically non-smooth and high frequency nature of action sequences, which poses a challenge for predictive modeling, as discussed in~\cite{tedrakeunderactuated}.
These methods have been effective within their respective domains.
Our model extends these concepts, incorporating deep fusion methods inspired by text-to-image applications, as showcased by~\cite{zhang2023adding}.

\section{Formulation and Method} \label{sec:formulation}
In this section, we reveal the formulation and technical details of the language-conditioned diffusion planner and the on-the-fly planning algorithm, which in combined is able to handle both fully and partially observable environments.

\subsection{Preliminary}
\mypar{\acf{ddm}} is a specific type of generative model that learns to model a data distribution by iteratively denoising from a pure noise~\cite{ho2020denoising} which eventually recovers the desired data (distribution).
Formally, a \ac{ddm} comprises two processes, \ie, a forward and a reverse diffusion process. The forward process is defined as: 

\begin{equation}
    q(\mathbf{x}_{1:T}|\mathbf{x}_0) = \prod_{t=1}^{T}q(\mathbf{x}_t|\mathbf{x}_{t-1})
    \label{eq:forward_diffusion}
\end{equation} 

\noindent where $\mathbf{x_t}$ is the noisy data at diffusion step $t$. $q(\mathbf{x}_t|\mathbf{x}_{t-1}) = \mathcal{N}(\mathbf{x}_t;\sqrt{1-\beta_t} \mathbf{x}_{t-1}, \beta_t \mathbf{I})$ is the Gaussian noise added to the data along the forward diffusion, and $\beta_t = 1-\alpha_t$ is the schedule. When $T \rightarrow \infty$, $\mathbf{x}_T$ is equivalent to an isotropic Gaussian distribution.

In order to generate a data point from a pure Gaussian noise $\mathbf{x_T} \sim \mathcal{N}(0, I)$, one needs to learn to model the noise in the forward process and revert the process (\autoref{eq:forward_diffusion}) by removing the learned noise.
Long story short, a parameterized probabilistic model $p_{\theta}$ is learned to infer the Gaussian noise. Specifically,

\begin{equation}
    p_{\theta}(\mathbf{x}_{t-1}|\mathbf{x}_{t}) = \mathcal{N}(\mathbf{x}_{t-1}; \mu_\theta(\mathbf{x}_t, t), \Sigma_\theta(\mathbf{x}_t, t))
    \label{eq:backward_diffusion}
\end{equation}

\noindent where the deviation $\Sigma$ is fixed and usually ignored in common practice. With the reparameterization trick, $\mathbf{x}_{t-1} = \frac{1}{\sqrt{\alpha_t}}(\mathbf{x_t}-\frac{\beta_t}{\sqrt{1-\bar{\alpha}_t}}\epsilon_\theta(\mathbf{x}_t, t))+\sqrt{\beta_t}\epsilon$, the objective of the model training is simplified as:

\begin{equation}
    L = || \epsilon_t - \epsilon_{\theta}(\mathbf{x}_{t}, t) ||^2
\end{equation}

\vspace{.3em}

\mypar{Notations.}
We hereby define the notations that would be used throughout the rest of the paper: $\mathbf{x}$ denotes the planning space; $o$ is a feature map transformed from agent observation; a predicted plan consists of trajectory $\tau$ and goal estimation $g$; where $t$ denotes the current planning time step. Meanwhile, $T$ represents the planning horizon. In order to eliminate the ambiguity, we use $k$ and $K$ to denote the current and the total number of diffusion steps going forward.

\subsection{Diffusion Model for Embodied Task Planning}
\label{sec:embodied_task_planning}
We use a \acf{ddm} to learn the task planning process of embodied AI. Formally, we define the task planning in embodied AI as a conditional probability of a state-sequence being generated: $p(\tau | o_t, I)$
\noindent where $\tau = (s_{t+1}, s_{t+2}, ..., s_{t+T})$ is the trajectory of the predicted states starting from the next time step, where the actual actions $a_t$ to be taken in the environment can be decoded using each consecutive predicted states ($s_t, s_{t+1}$).
We defer the detailed discussion of the action decoding to~\secref{sec:learning_and_planning}.
$I$ here denotes the language instruction.

\begin{figure}[t]
    \centering
    \includegraphics[width=0.85\linewidth]{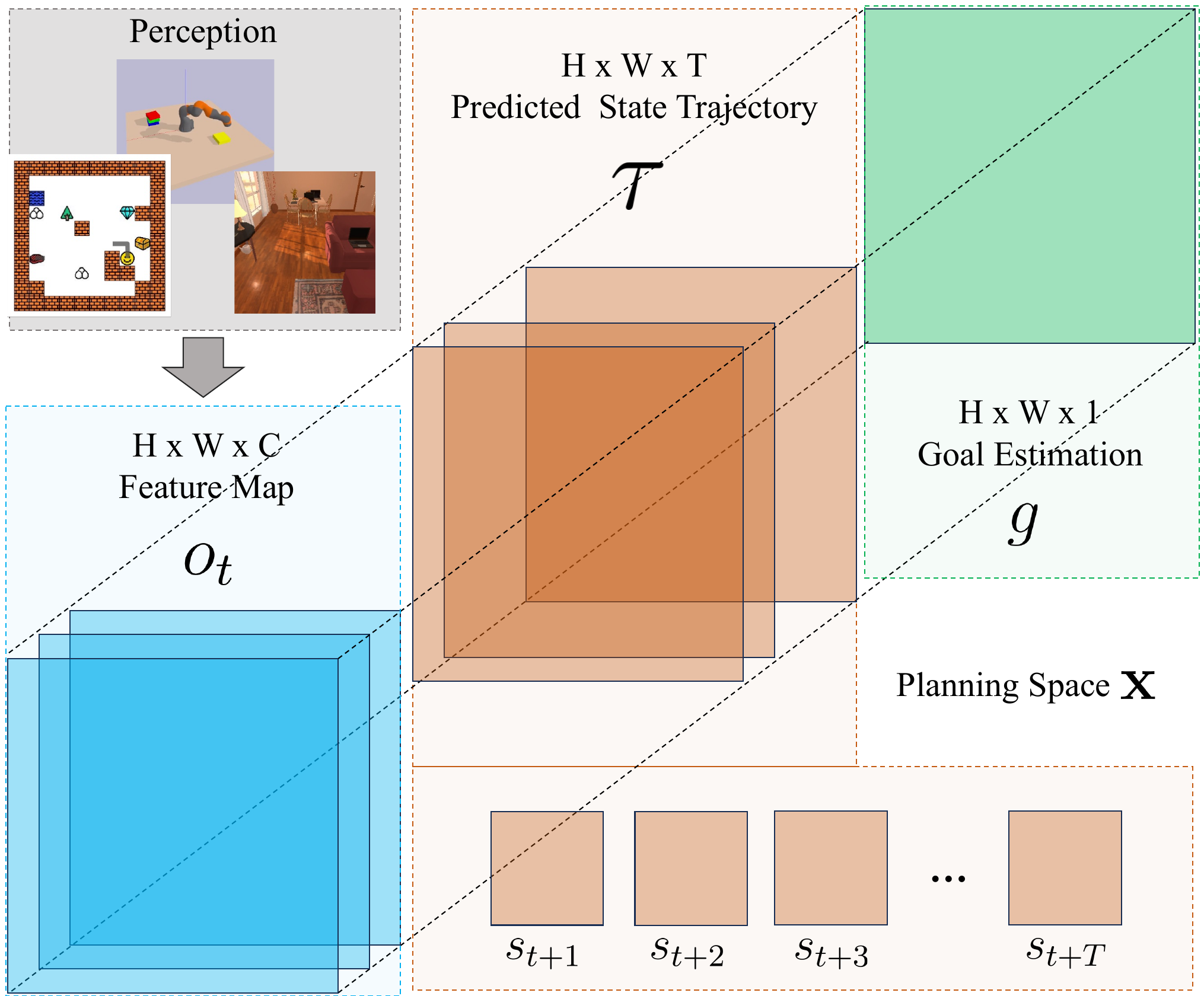}
    \caption{Planning space of ``Planning as in-painting''}
    \label{fig:planning_space}
    \vspace{-.6em}
\end{figure}

\vspace{.3em}

\begin{figure*}[t]
    \centering
    \includegraphics[width=\linewidth]{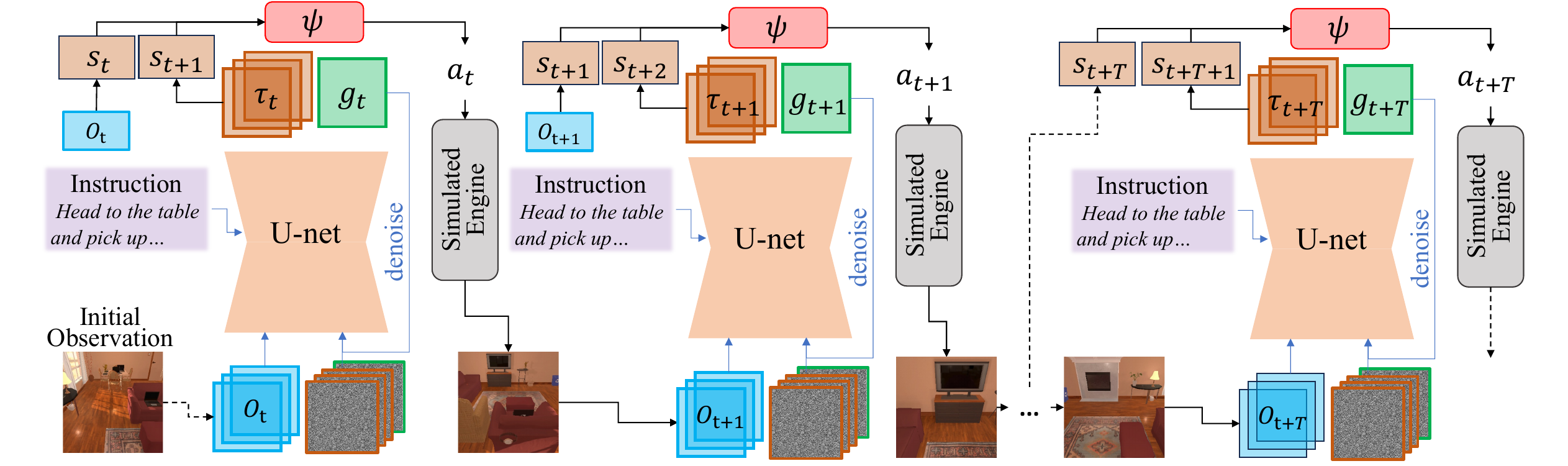}
    \caption{Illustration of the proposed framework: integrating the diffusion-based planner into the on-the-fly algorithm.}
    \label{fig:on_the_fly}
\end{figure*}

\mypar{Planning Space.} 
In our method, we define the planning space as $\mathbf{x} \in \mathcal{R}^{H \times W \times (C+T)}$, where $H \times W$ is the dimension of the environment observation, $C$ represents the channels of environmental information at the current time step, and $T$ encapsulates the temporal dimension of the plan.
Specifically, our model conditions on the observed environment $o_t$, which is a task feature map $o_t \in \mathcal{R}^{H \times W \times C}$ derived from perceptual input and includes all pertinent environmental data such as object locations and the state of interactive elements---and predicts $\tau$, which is a sequence of $T$ future states $\tau \in \mathcal{R}^{H \times W \times T}$.
Each state map, $s_t \in \mathcal{R}^{H \times W}$, represents a predicted concrete status at time step $t$. 
For example, in navigation tasks, $s_t$ might indicate the agent's projected location, whereas in object manipulation, it could denote the expected position of a target object (and hence drives how a robotic arm should approach it).
By concatenating these state maps, we obtain a trajectory $\tau$ that spans the planning horizon.
This formulation, while exemplified here with navigation and object manipulation, is designed to be adaptable to a variety of embodied visual task planning problems.


\vspace{.3em}

\mypar{Planning in Partially Observable Environment.}
In some of the embodied AI tasks, the agent is not exposed to the entire environment. In other words, the environment observation $o$ is partially observable. Thus the predicted plan tends to be unreliable with insufficient information~\cite{ortega2021shaking}.
In order to collect sufficient information for a more accurate plan, the planning algorithm of an embodied AI often relies on exploratory probing to the environment, \eg, random walking to \textit{explore} an unseen environment, or attempting manipulating an occluded object in a trial and error manner.
However, the stochacity of such exploratory phase hinders an efficient (and sometimes effective) solution of the plan and may suffer from catastrophic failures when the plan is deployed in the wild.
To alleviate such an issue, we introduce another variable $g \in \mathcal{R}^{H\times W}$ to estimate the goal state in the prediction subspace in $\mathbf{x}$ of the task planning and reformulate the conditional state-sequence probability into:

\begin{equation}
    p(\tau, g | o_t, I) = p(\tau | g, o_t, I) p(g | o_t, I)
\end{equation} \label{eq:task_planning_partial}

We predict the joint distribution of the plan $\tau$ together with an estimation of the goal $g$ given the current observation $o_t$ and language instruction $I$. The planning space under partially observable environment is thus extended to $\mathbf{x} \in \mathcal{R}^{H \times W \times (C+T+1)}$. The joint probability of $\tau$ and $g$ can be rewritten by probability chain rule into \autoref{eq:task_planning_partial}. The right-hand side is actually a general formulation of decision making with partially observable environment. It provides a succinct theoretical justification to answer why the estimated goal $g$ could help improving the accuracy of the plan $\tau$. Our experiments also show the effectiveness of incorporating $g$ into the formulation.

\vspace{.3em}

\mypar{Planning as ``In-Painting".}
As previously mentioned, we adopt the \ac{ddm} as the planner for embodied task planning. However, the common practice to leverage \ac{ddm} for data generation is from pure Gaussian noise to the recovered data. In our case, the condition $o_t$ is provided by the perception modules and it cannot be modified during one step planning. Therefore, prediction $\tau$ of a plan $\mathbf{x}$ is the part that needs to be generated by the \ac{ddm}. The formulation of generating the trajectory $\tau$ and goal estimation $g$ in the subspace of the plan $\mathbf{x}$ is very similar to image in-painting given an incomplete image. Formally, we perform reverse diffusion process on $\tau$ and $g$ which are the $C+1$ to $C+T$ and $C+T+1$ channels in $\mathbf{x}$. These channel will be initialized as Gaussian noise at the beginning of each planning step. The predicted trajectory $\tau$ and goal estimation $g$ will be generated after iterating $k$ step of denoising process by the diffusion algorithm.

\subsection{Learning and Planning}
\label{sec:learning_and_planning}

\mypar{Training data preparation.}
To train our diffusion-based planner, we collect the data from three most popular families of embodied tasks: (i) navigation, (ii) object manipulation, (iii) task planning in interactive embodied environment. All collected data will be transformed into task feature maps. Each data contains four key components (1) a ground-truth task planning trajectories $\tau_{gt} \in \mathcal{R}^{H \times W \times T}$ in the semantic space together with (2) the accompanied set of language instructions $I$, (3) the task feature maps transformed from sequential observations $o_{0:T} \in \mathcal{R}^{H \times W \times C}$ along each ground-truth plan, and (4) the ground-truth goal $g_{gt}$. Although the dimension of the maps varies from task to task, it won't affect the training and planning.

\vspace{.3em}

\mypar{Learning.} We follow the supervised training paradigm used in \ac{ddm}~\cite{ho2020denoising}. Concretely, each trajectory $\tau = (s_{t+1}, s_{t+2}, ..., s_{t+T}), \tau \in \mathcal{R}^{H \times W \times T}$ within the environment is paired with a corresponding task instruction $I$, task feature maps $o = (o_{t+1}, o_{t+2}, ..., o_{t+T}), o_t \in \mathcal{R}^{H \times W \times C}$, and goal state estimation $g \in \mathcal{R}^{H \times W}$. During training, we commence by sampling noise $\epsilon \sim \mathcal{N}(0,1), \epsilon \in \mathcal{R}^{H \times W \times T}$ and a specific time step $k \sim \mathcal{U}(0,K)$. For simplicity, we rewrite $\mathbf{w} = (\tau, g)$. Subsequently, we construct noisy trajectories $x_k(\mathbf{w}) = \sqrt{\alpha_k}\mathbf{w} + \sqrt{1-\alpha_k}\epsilon$ by systematically injecting noise, to introduce stochasticity that is essential for the diffusion model to learn the underlying denoising process. Our noise prediction model $\epsilon_\theta$ is jointly conditioned on noisy input $x_k(\mathbf{w})$, task feature map $o$ and the instruction $I$ to predict the noise. By minimizing the discrepancy between the ground-truth noise $\epsilon$ and the predicted one, our model learns to reverse the diffusion process. Thus, our training objective can be written as:

\begin{equation}
    \mathcal{L}(\theta) = \mathbb{E}_{q(\mathbf{w}_{1:K}, \boldsymbol{\epsilon}_{1:K}|\mathbf{w}_0)}\left[\sum_{k=1}^{K} \left\| \boldsymbol{\epsilon}_k - \epsilon_{\theta}(x_k(\mathbf{w}), k, o, I) \right\|^2 \right]
\end{equation}  \label{eq:learning_objective}

Nevertheless, unlike the common practice in conditional image generation, we choose not to take advantage of \textit{Classifier Free Guidance (CFG)} for learning. Due to the nature of instruction-following, both the predicted trajectory and the goal estimation have to be conditioned on the instructions; otherwise, they are meaningless. We also found significant performance drop while using CFG.

\vspace{.3em}

\mypar{Planning.}
In most of the time, the goal estimation $g$ could improve the accuracy of the generated plan $\tau$. However, both of them are unlikely to be accurate if the observation only contains limited information of the environment, \eg, when the agent is at the beginning phase of the task. In order to improve the success rate of the final plan execution, we design an on-the-fly planning algorithm as shown in Algorithm \autoref{alg:on_the_fly}.

We formal version of the on-the-fly planning algorithm is illustrated in Algorithm \autoref{alg:on_the_fly}. The key idea of Algorithm \autoref{alg:on_the_fly} is to improve the accuracy of the generated plan by discovering more meaningful information along the way of plan execution. Comparing to random exploration, executing the front part of trajectory $\tau=s_{t+1:t+T}$ should be more efficient because the diffusion model learns from ground-truth plans. After taking one step $s_{t+1}$ in the generated trajectory $\tau$ at each time $t$, the agent will reveal more information in the environment. As the exploration going, the diffusion model keeps updating the plan, \ie trajectory $\tau$ and goal estimation $g$, with more complete environmental information. As a result, it could increase the final execution successful rate. Our experimental results prove the effectiveness of Algorithm \autoref{alg:on_the_fly}.

\vspace{.3em}

\begin{algorithm}[t]
    \begin{algorithmic}[1]
        \Inputs{Planning Horizon: $T$, Instruction: $I$}
        \Initialize{$o_t \gets o_0, s_t \gets s_0, t \gets 0$}
        \For{t = 0 to T}
            \State $\tau_K \sim \mathcal{N}(0, I)$, $b_K \sim \mathcal{N}(0, I)$
            \State $\mathbf{w}_K = (\tau_K, b_K)$
            \For{k = K to 1}
                \State $z \sim \mathcal{N}(0, I)$ if $k>1$, else $z=0$
                \State $\mathbf{w}_{k-1} \gets \frac{1}{\sqrt{\alpha_k}}(\mathbf{w}_{k} - \frac{1-\alpha_k}{\sqrt{1-\bar{\alpha}_k}} \epsilon_\theta(\mathbf{w}_{k}, k)) + \sigma_k z$
            \EndFor
            \State $\tau, g = \mathbf{w}_0$
            \State $s_{t+1}, ..., s_{t+T} \gets \tau$
            \State $a_t \gets \psi(s_t, s_{t+1})$
            \State $o_{t+1} \gets Env(s_t, a_t)$
            \State $t \gets t+1$, $o_t \gets o_{t+1}$, $s_t \gets s_{t+1}$
            \If {$o_t$ is terminated}
                \State break
            \EndIf
        \EndFor
    \end{algorithmic}
    \caption{On-the-fly planning with diffusion planner.}
    \label{alg:on_the_fly}
\end{algorithm} 

\mypar{Execution.}
The generated trajectory $\tau=s_{t+1:t+T}$ represents a sequential states of the agent in the environment. It is not directly executable because actions are not modeled. Fortunately, it actually brings more flexibility for plan execution. We can use different motion planning algorithms accordingly to implement the state transition between two adjacent time steps. Formally, we model the deterministic state transition with an inverse dynamics model $\psi(s_t, s_{t+1})$ that generates an action for plan execution.

\begin{equation}
    a_t = \psi(s_t, s_{t+1})
\end{equation}

Note that $a_t$ is not necessary to be one atomic action in the legal action space. In other words, $a_t$ could also represent a sequence of legal actions. The inverse dynamics model varies in different tasks and environments. It could be an A* algorithm in discrete space navigation. For object manipulation, it can be an inverse kinematics motion planner.
\section{Experiments} \label{sec:experiment}

Our experiments are designed to answer the following research questions:
(i) Will the proposed method of "planning as in-painting" function as an effective substitute for RL policies when it comes to goal-driven planning tasks?
(ii) Can the proposed technique be seamlessly integrated or adapted to a diverse set of tasks?
(iii) Can diffusion-based planning models incorporate language instructions more effectively than conventional RL algorithms?

Specifically, to answer the aforementioned question (ii), we evaluate the proposed framework in three environments with increasing planning complexities:
(1) \textbf{CompILE}~\cite{kipf2019compositional}: a \textbf{multi-goal grid world} with a series of goal-objects instructed,
(2) \textbf{Kuka Robot}~\cite{schreiber2010fast}: a \textbf{block-stacking and rearrangement} environment where a robotic arm follows a language-specified target stacking setup,
and (3) \textbf{ALFRED}~\cite{shridhar2020alfred}: a complex indoor environment with \textbf{diverse objects and tasks} where an agent needs to traverse and interact with the environment following given instructions.

For each environment, we answer the question (i) by comparing our proposed algorithm with carefully designed and trained RL policies. 
Additional ablation studies are subsequently conducted to examine the effectiveness of integrating language instructions into planning modules in response to question (iii).
In the following sections, we provide detailed explanations of the configurations for each environment, the corresponding evaluation metrics, the baseline models to compare with, and the experimental results of the proposed method.
Model architecture and hyper-parameters are in the appendix.

\subsection{Multi-Goal Grid World: CompILE}
\label{sec:compile}
\mypar{Settings. }
We first consider a synthetically built grid-world environment, CompILE~\cite{kipf2019compositional}.
To further increase the problem's complexity, we add five more different objects and augment the map size to $16 \times 16$.
The environment contains at most $10$ obstacles that the agent cannot step on.
In this grid map, an agent is required to navigate to multiple targets according to a natural language instruction.
The episode length is set to be $100$ steps.
For each episode, the target objects are randomly initialized.

For quantitative analysis, we examine three different settings with increasing complexities: 1. One object and one goal (\textbf{1O1G}): the environment contains a single object positioned at a random initial location for the agent to reach.
2. Multi-object and one goal (\textbf{MO1G}): the environment contains up to $8$ objects, and the agent's objective is to navigate to the object indicated by the instruction. 
3. Partially observable (\textbf{P-MO2G}) setting: the target object is not visible to the agent at the beginning; however, the instruction will provide guidance on a reference object that the agent can first navigate to. This setup evaluates the model's capacity to deduce the accurate goal state and navigate to a target that is not visible initially. For example, \textit{'First, go to the apple, then turn right and head towards the tomato'}, where the tomato is not initially visible (to the agent's FoV).

\mypar{Evaluation Metrics.} We employ several metrics for evaluation: task success rate (SR); and goal distance (GD), measuring the Euclidean distance from the agent's final position to the target. Inspired by~\cite{jain2019stay}, to test if the agent has faithfully followed the instructions, we define the overlap between the predicted and ground-truth paths as true positives and the missed ground-truth path points as false negatives. False positives are defined as non-path predictions that are incorrectly marked as the path.

\mypar{Implementation Details.}
As mentioned in~\secref{sec:embodied_task_planning}, our planning space $\mathbf{x}$ is in $\mathcal{R}^{H \times W \times (C+T+1)}$, where the predicted future states (\ie, the plan) are in $\mathcal{R}^{H \times W \times T}$ and the predicted goal estimation is in $\mathcal{R}^{H \times W}$. In this task, we set $C = 512$ where the task feature map is the BERT~\cite{devlin2018bert} representation of the object that occupies the grid. Due to page limits, we state the rest of the details in the appendix.


\mypar{Baselines.}
We benchmark the proposed method against three key baselines: Conservative Q-Learning (CQL) \cite{NEURIPS2020_0d2b2061} (the state-of-the-art model-free offline \acf{rl} method), behavioral cloning (BC), and the cutting-edge diffusion policy Decision Diffuser (denoted as Dec-Diffuser)~\cite{ajay2022conditional}.
Both CQL and BC are adapted in our implementation where we use one-hot vector to represent the goal. This is necessary as the goal can correspond to any object within the environment, the one-hot vector here mirrors the concept of language guidance. The reward function is implemented as the distance from the agent to the goal. The architecture of CQL and BC models consists of two layers of convolution neural networks followed by two fully connected layers. The one-hot vector for goal guidance is introduced by concatenating with the output of the model just prior to the final fully connected layer. For behavioral cloning, we generated the desired path using the A* search algorithm. On the other hand, we adapt the Decision Diffuser (DD) into our setting by flattening the grid into a $256$-dimensional vector and training the DD to predict the entire trajectories generated by the A* algorithm.

\definecolor{LightCyan}{rgb}{0.88,1,1}
\newcolumntype{a}{>{\columncolor{LightCyan}}d}

\definecolor{LightGreen}{rgb}{0.9,1,0.9}
\newcolumntype{a}{>{\columncolor{LightGreen}}c}

\begin{table}[tp!]
\centering
\small\addtolength{\tabcolsep}{-1.4pt}
\scalebox{.86}{
    \begin{tabular}{cc|ccccc}
    \toprule
    \multicolumn{1}{c}{\multirow{2}{*}{\textbf{Setting}}} & \multicolumn{1}{c|}{\multirow{2}{*}{\textbf{Models}}} & \multicolumn{5}{c}{\multirow{1}{*}{\textbf{Metrics}}}
    \\
    & &
    SR & GD & Path Prec. & Path Rec. & \multicolumn{1}{c}{\multirow{1}{*}{Path F1}} \\
    \midrule

    \multirow{4}{*}{1O1G}
    & \multirow{1}{*}{BC}
      & 100 & 0.00 & 5.13 & 36.87 & 9.01
    \\
    & \multirow{1}{*}{CQL}
      & 83.80 & 0.44 & 5.1 & 49.65 & 9.25
    \\
    & \multirow{1}{*}{DD}
      & 0.003 & 5.92 & - & - & -
    \\
    & \multirow{1}{*}{\textbf{Ours}}
      & 99.9 & 0.002 & 99.97 & 99.95 & 99.96
    \\
      
    \midrule

    \multirow{4}{*}{MO1G}
    & \multirow{1}{*}{BC}
      & 23.20 & 6.49 & 4.42 & 66.55 & 8.29
    \\
    & \multirow{1}{*}{CQL}
      & 3.03 & 9.06 & 4.75 & 48.82 & 8.65
    \\
    & \multirow{1}{*}{DD}
      & - & - & - & - & -
    \\
    & \multirow{1}{*}{\textbf{Ours}}
      & 99.50 & 0.03 & 99.63 & 99.42 & 99.52
    \\
    
    \midrule

    \multirow{4}{*}{P-MO2G}
    & \multirow{1}{*}{BC}
      & 19.19 & 8.27 & 4.84 & 74.09 & 82.61
    \\
    & \multirow{1}{*}{CQL}
      & 3.03 & 9.77 & 4.87 & 80.93 & 86.69
    \\
    & \multirow{1}{*}{DD}
      & - & - & - & - & -
    \\
    & \multirow{1}{*}{\textbf{Ours}}
      & 99.9 & 0.002 & 93.34 & 96.48 & 94.88
    \\

    \bottomrule
    
    \end{tabular}
}
\caption{
Experimental results of our method on CompILE. For detailed settings, please refer to Sec. \ref{sec:compile}.
}
\label{tab:compile_results}
\end{table}
\mypar{Results.}
The performances of both our method and the baselines are shown in \autoref{tab:compile_results}. In the simplest setting, \ie \textbf{1O1G}, both our method and BC achieve almost perfect results; CQL also shows reasonable performance, but DD performs surprisingly badly. The inductive bias (a 1-D CNN) used by DD may heavily impact the performance as it doesn't model spatial correlations. As there are more objects as in \textbf{MO1G}, neither BC nor CQL can plan reliably anymore. In partially observable settings (\textbf{P-MO2G}), their performances are even worse and tend to make random guesses. Our method, however, demonstrates strong reliability and robustness in both fully and partially observable settings, which justifies the design of our modeling and planning algorithms.

\subsection{Robotic Arm Block Stacking: Kuka} \label{sec:kuka}

\mypar{Settings.}
To investigate the efficacy of our ``planning as in-painting'' framework in manipulation tasks with complex visual observations, we conduct additional experiments in the Kuka environment~\cite{schreiber2010fast} following~\cite{ajay2022conditional}.
Here, a robotic arm manipulates a few blocks and constructs them into a goal configuration following language instructions. Our method generally predicts the end-effector of the arm, and we apply an inverse kinematic control to move the robotic arm to our predicted desired end state. The stacking process iterates until it reaches the maximum episode length or the goal configuration.
The following two tasks are performed. 1. \textbf{Stacking}: assemble a tower consisting of four blocks. 2. \textbf{Rearrangement}: place one set of blocks on top of another. 
Additional constraints that specify the spatial relationship of the blocks are applied as well. The configurations may result in multiple structures, such as a single tower and two separate ones, presenting an out-of-distribution challenge for the model to interpret/execute the stack-place relations.

\vspace{.3em}

\mypar{Implementation Details and Baselines.} Similarly, we compare our methods with CQL, BC, Diffuser~\cite{janner2022planning}, and Decision Diffuser~\cite{ajay2022conditional} (DD). For Diffuser and DD, we directly report the number from their paper. For CQL and BC, we use one-hot vector to denote the constraint, and for our own methods, we use instructions to represent the constraints, and the dimension of our task feature map $o \in \mathcal{R}^{H \times W \times C}, C = 1$ represents the position of each block in the environment.

\mypar{Evaluation Metrics.} For Kuka dataset, we consider task success rate as our metric. Specifically, the task success rate is defined as the number of satisfied constraints divided by the number of specified constraints.

\vspace{.3em}

\definecolor{LightCyan}{rgb}{0.88,1,1}
\newcolumntype{a}{>{\columncolor{LightCyan}}d}

\definecolor{LightGreen}{rgb}{0.9,1,0.9}
\newcolumntype{a}{>{\columncolor{LightGreen}}c}

\begin{table}[tp!]
\centering
\small\addtolength{\tabcolsep}{-1.4pt}
\scalebox{1}{
    \begin{tabular}{c|cccc}
    \toprule
    \multicolumn{1}{c|}{\multirow{2}{*}{\textbf{Models}}} & \multicolumn{2}{c}{\multirow{1}{*}{\textbf{Single Constraint}}} & \multicolumn{2}{c}{\multirow{1}{*}{\textbf{Multi Constraint}}}
    \\
    & Stack & \multicolumn{1}{c}{\multirow{1}{*}{Rearrange}} & Stack & Rearrange \\
    \midrule

    \multirow{1}{*}{BC}
      & 4.00 & 20.00 & 0.0 & 0.0
    \\
    \multirow{1}{*}{CQL}
      & 0.0 & 5.00 & 0.0 & 0.0
    \\
    \multirow{1}{*}{Diffuser}
      & 45.60 & 58.90 & - & -
    \\
    \multirow{1}{*}{DD}
      & 58.00 & 62.70 & 60.30 & 67.20
    \\
    \multirow{1}{*}{\textbf{Ours}}
      & 99.50 & 99.80 & 98.38 & 98.85
    \\
    

    \bottomrule
    
    \end{tabular}
}
\caption{
The experimental results of our method on Kuka object manipulation. For detailed settings, please refer to Sec. \ref{sec:kuka}.}
\label{tab:kuka_results}
\end{table}
\mypar{Results.}
From \autoref{tab:kuka_results}, we find that both BC and CQL fail in this task because they are goal-specific. In other words, when there are different goals, they are not able to associate them with the planned trajectories. Both our method and DD, in contrast, show much better generalizability in all settings. However, DD still falls behind by a large gap because it doesn't explicitly model language instructions, which are pivotal for generalizing across different task constraints, thereby enabling our method to outperform other baselines by a visible margin.

\subsection{ALFRED}

\mypar{Settings.} 
To evaluate our framework's efficacy in realistic, complex settings, we utilize ALFRED~\cite{shridhar2020alfred}—a simulated household environment for instruction-following tasks. Here, an agent must comprehend language directives to locate and interact with specific objects. Each directive comprises two components: a navigation instruction to guide the agent towards the target, and a subsequent command specifying the object for interaction. Given that the target object is typically not visible from the starting position, the agent is required to infer from instruction cues to successfully locate and interact with target object. We assessed our methods and baselines on the validation set, distinguishing between \textit{seen} and \textit{unseen} environments. The \textit{seen} category refers to environments that were included in the training data, whereas the \textit{unseen} are those that the agent was not exposed to during training. Our evaluation on the \textit{unseen} split offers a direct measure of our method's ability to generalize to new environments not encountered during training.

\begin{table}[tp!]
    \centering
    \scalebox{0.92}{
        \begin{tabular}{c|c cc c}
            \toprule
             \multicolumn{1}{c|}{\multirow{2}{*}{\textbf{Models}}}  & \multicolumn{2}{c}{\textbf{Seen}} &  \multicolumn{2}{c}{\textbf{Unseen}}      \\
                                & SR         & GC          & SR              & GC            \\
             \midrule
             EmBERT             & \textbf{37.4/28.8}  & \textbf{44.6/36.4}   & 5.7/3.1         & 15.9/9.3      \\
             E.T                & 34.7/24.6  & 42.0/31.0   & 3.5/1.8         & 13.6/8.0      \\
             LACMA              & 36.9/27.5  & 42.8/33.8   & 8.2/5.1         & 18.0/12.2     \\
             FILM               & 25.4/10.3  & 38.8/14.8   & 20.4/8.1        & 32.7/12.2     \\
             \midrule
             \textbf{Ours}      & 31.2/15.3  & 42.0/17.8   & \textbf{26.1/10.0}       & \textbf{39.7/12.5}     \\
            \bottomrule
        \end{tabular}
        }
    \caption{Experimental results of our method on ALFRED. The path-length weighted results are presented following the slash.}
    \label{tab:alfred}
\end{table}
\vspace{.3em}

\mypar{Implementation Details.} For this task, we build our model upon FILM~\cite{min2021film}. Following~\cite{min2021film, blukis2022persistent}, we use a semantic mapping module to project the field of view images of the agent into a 3D voxel, then, we do a max-pooling along the z-dimension to get a 2D semantic map, which serves as the task feature map $o$ in our method.

\vspace{.3em}

\mypar{Baselines.} For baselines, we consider end-to-end methods Embodied BERT~\cite{suglia2021embodied}, Episodic Transformer~\cite{pashevich2021episodic}, and LACMA~\cite{yang2023lacma}, which demonstrate great performance in the~\textit{seen} environment. Meanwhile, we consider modular methods which show great generalizability in the~\textit{unseen} environment.

\mypar{Evaluation Metrics.} We consider task success rates (SR), goal-conditioned success rates (GC) and the path-length-weighted (PLW) for the metrics. SR quantifies the proportion of tasks in which the agent successfully completes all subgoals, whereas GC calculates the fraction of subgoals achieved at the end of the task, and PLW adjusts the success rate by the efficiency of the path taken, reflecting a balance between accuracy and task completion efficiency.

\mypar{Results.} 
From~\autoref{tab:alfred} we can see that our model significantly improves the performance on the~\textit{unseen} split by $5.7\%$ comparing to the baseline FILM that our model builds upon. This verifies our assumption that our model is able to comprehend the complex instructions under realistic setting, and thus finding the target objects even if the model has never been trained on the environment before.

\subsection{Ablation Study} We present the performance of the ablated versions of our model on the CompILE dataset in~\autoref{tab:ablation}. Our baseline model uses a 1 channel grid to reprsent the initial states and we only perform one-shot prediction during inference time. While the second row means our model is conditioned on the task feature map which use BERT embedding to represent object in each grid. From the table one can see the contextualize embedding of the grid provides more information to our model, resulting in slight improve under the MO1G setting. While the goal estimation $g$ is essential to our method especially when the environment is partially observable, it not only guides our model to find the target under the fully observable setting, it also boosts our model with a significant improvement under the partially observable setting. In addition, we find that under the partially observable setting, it's important to continually update the environment and plan accordingly, resulting in the best results in the table.

\definecolor{LightCyan}{rgb}{0.88,1,1}
\newcolumntype{a}{>{\columncolor{LightCyan}}d}

\definecolor{LightGreen}{rgb}{0.9,1,0.9}
\newcolumntype{a}{>{\columncolor{LightGreen}}c}

\begin{table}[tp!]
\centering
\small\addtolength{\tabcolsep}{-1.4pt}
\scalebox{.80}{
    \begin{tabular}{cc|ccccc}
    \toprule
    \multicolumn{1}{c}{\multirow{2}{*}{\textbf{Setting}}} & \multicolumn{1}{c|}{\multirow{2}{*}{\textbf{Models}}} & \multicolumn{5}{c}{\multirow{1}{*}{\textbf{Metrics}}}
    \\
    & &
    SR & GD & Path Prec. & Path Rec. & \multicolumn{1}{c}{\multirow{1}{*}{Path F1}} \\
    \midrule

    \multirow{4}{*}{MO1G}
    & \multirow{1}{*}{baseline}
      & 97.10 & 0.25 & 98.20 & 96.50 & 97.34
    \\
    & \multirow{1}{*}{$+$ grid repr}
      & 97.40 & 0.23 & 99.70 & 96.64 & 98.15
    \\
    & \multirow{1}{*}{$+ g$}
      & 99.50 & 0.002 & 99.63 & 99.42 & 99.52
    \\
    & \multirow{1}{*}{$+$ on the fly}
      & - & - & - & - & -
    \\
    
    \midrule

    \multirow{4}{*}{P-MO2G}
    & \multirow{1}{*}{baseline}
      & - & - & - & - & -
    \\
    & \multirow{1}{*}{$+$ grid repr}
      & 6.90 & 6.38 & 21.95 & 20.64 & 21.27
    \\
    & \multirow{1}{*}{$+ g$}
      & 28.60 & 4.05 & 31.07 & 33.16 & 32.08
    \\
    & \multirow{1}{*}{$+$ on the fly}
      & 99.89 & 0.002 & 93.34 & 96.48 & 94.88
    \\
    \bottomrule
    \end{tabular}
}
\caption{
Ablation study results of our method on CompILE.
}
\label{tab:ablation}
\end{table}

\subsection{Additional Analysis}
\label{subsec:analysis}
\begin{figure}
    \centering
    \includegraphics[width=\linewidth]{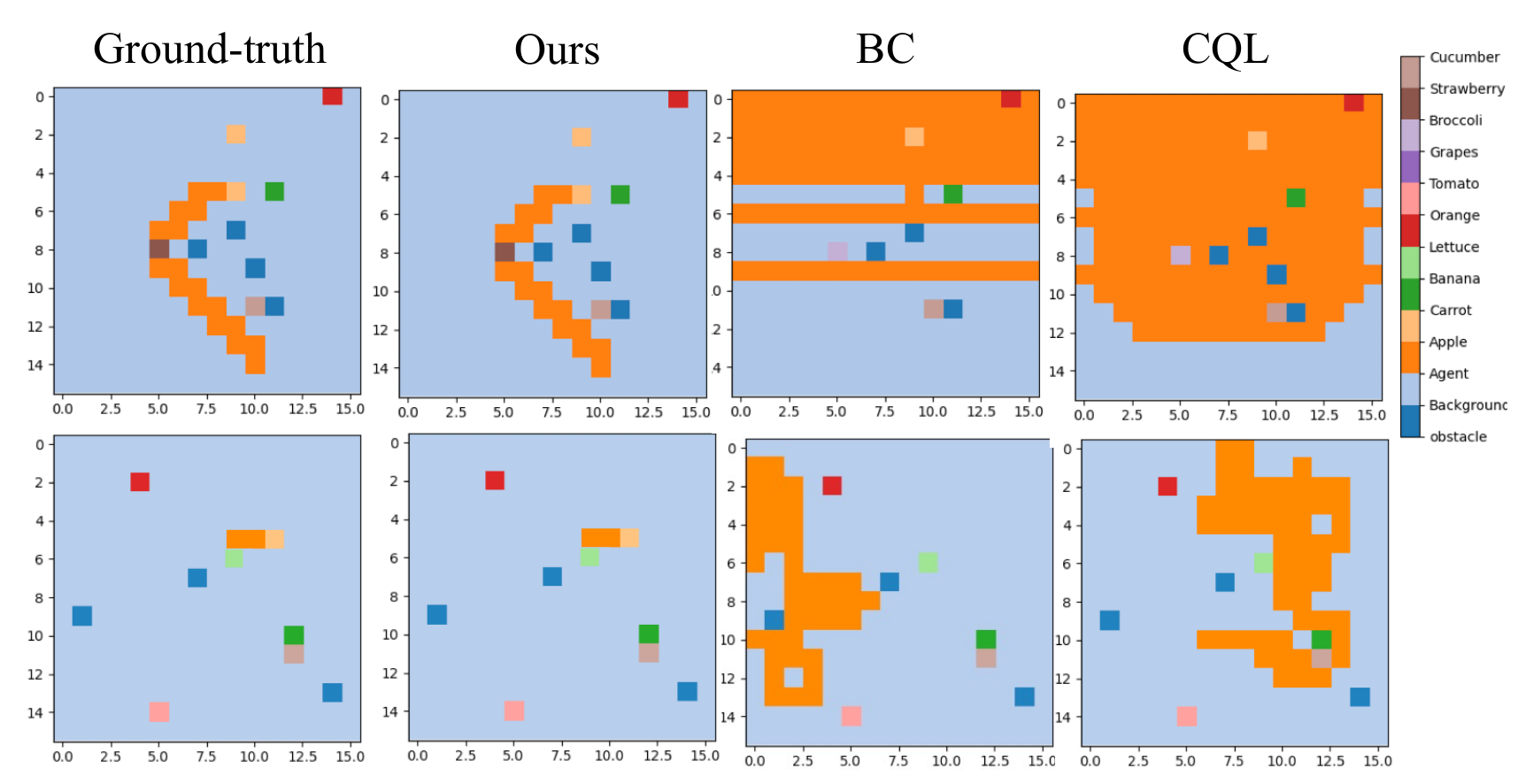}
    \caption{Qualitative Results on CompILE.}
    \label{fig:qualitative}
\end{figure}

\mypar{Qualitative Analysis.} In our qualitative analysis, experiments were carried out on the CompILE dataset, where we illustrated the predicted trajectories for each method. As evidenced in~\autoref{fig:qualitative}, one can see that our methods predict the path that head straightforward to the object, while CQL and BC require exploration, which is less efficient.

\definecolor{LightCyan}{rgb}{0.88,1,1}
\newcolumntype{a}{>{\columncolor{LightCyan}}d}

\definecolor{LightGreen}{rgb}{0.9,1,0.9}
\newcolumntype{a}{>{\columncolor{LightGreen}}c}

\begin{table}[tp!]
\centering
\small\addtolength{\tabcolsep}{-1.4pt}
\scalebox{1}{
    \begin{tabular}{c|cc|cc}
    \toprule
    \multicolumn{1}{c|}{\multirow{2}{*}{\textbf{Models}}} & \multicolumn{2}{c|}{\multirow{1}{*}{\textbf{Single Constraint}}} & \multicolumn{2}{c}{\multirow{1}{*}{\textbf{Multi Constraint}}}
    \\
    & Stack & \multicolumn{1}{c|}{\multirow{1}{*}{Rearrange}} & Stack & Rearrange \\
    \midrule

    \multirow{1}{*}{Orig LI}
      & 99.50 & 99.80 & 98.38 & 98.85
    \\
    \multirow{1}{*}{Complex LI}
      & 98.70 & 97.80 & 97.33 & 78.55
    \\
    

    \bottomrule
    
    \end{tabular}
}
\caption{Performances of the model trained on different language instruction complexity.}
\label{tab:instructions}
\vspace{-1.em}
\end{table}
\vspace{9pt}
\mypar{Complexity of Language Instructions.}
To assess our model's ability to interpret varied instructions, we carried out experiments on the Kuka dataset, training our model with instructions that direct to the same goal but are phrased differently. The outcomes are detailed in \autoref{tab:instructions}. We utilized ChatGPT to rephrase the original directives and intentionally altered spatial prepositions, such as changing \textit{"on"} to \textit{"below"}. The precise instructions used are listed in the appendix. The results demonstrate that our model maintains performance comparable to the original baseline, even when processing more complex language variations, confirming its robustness in incorporating linguistic cues.

\section{Conclusions}

In this work, we present a novel embodied AI planning framework named ``planning as in-painting'' as well as an on-the-fly algorithm which is based on a language-conditioned diffusion model. It can infer instruction-intended goal and plan optimal robotic trajectories in environments under uncertainty. We demonstrated the effectiveness of our method in three tasks across different domains. 
We envision that our method can be extended to model 3D agent dynamics.

\newpage
{
    \small
    \bibliographystyle{ieeenat_fullname}
    \bibliography{references}
}

\clearpage
\setcounter{page}{1}
\setcounter{section}{0} 
\renewcommand{\thesection}{\Alph{section}} 

\maketitlesupplementary

\section{Implementation Details}
\label{supp:implement}

Generally, we use \texttt{UNet2DConditionModel} as the backbone of our diffusion model. For training, we set the batch size to $64$ and the total training epoch to be $200$. Our learning rate begins at $0$ and linearly increases to $0.0002$ over the first 5 epochs, after which it undergoes a cosine decay back to $0$. For noise scheduling strategies, we utilize DDIM~\cite{song2020denoising} and a cosine beta schedule where the noise level follows a squared cosine function with a capping mechanism. For both training and inference, we set the diffusion step to 100. We now detail task-specific parameters and implementations.

\subsection{Multi-Goal Grid World: CompILE}

\mypar{Ours.} For one object and one goal (\textbf{1O1G}), our UNet has $3$ input channels and $2$ output channels, where the first input channel is the initial grid, and the second and thrid channels are noisy channels as described in~\autoref{sec:learning_and_planning}. The first output channel depicts the path, and the second channel is the goal estimation. Specifically, the UNet contains $6$ blocks, for downsample blocks, we have (\texttt{DownBlock2D, AttnDownBlock2D, AttnDownBlock2D}), and for upsample blocks, we have~(\texttt{AttnUpBlock2D, AttnUpBlock2D, UpBlock2D}). The output channel for the blocks are ($8, 16, 64$) respectively.
 
For multi-object and one goal (\textbf{MO1G}), since the environment is much more complex, we first convert the $16\times16$ grid into a representation grid with the shape of $16\times16\times512$. Here, each grid cell contains a 512-dimensional textual representation of the object occupying that cell. We use~\texttt{CLIPTextModel} to extract the word embedding of the object. The input channel is $514$, where the first $512$ channels are the representation of the grid, and the last two are noisy channels. The output channel of the model is $2$, which also consists of the predicted path and estimated goal. Similarly, the UNet contains $6$ blocks, for downsample blocks, we have (\texttt{CrossAttnDownBlock2D, CrossAttnDownBlock2D, DownBlock2D}), and for upsample blocks, we have~(\texttt{UpBlock2D, CrossAttnUpBlock2D, CrossAttnUpBlock2D}). The output channels for the blocks are ($256, 512, 1024$) respectively. Since there are multiple objects in the environment, we use textual instructions to specify the goal, which is later transformed into textual representations with the shape of ($L\times512$), where $L$ is the length of the instruction. To fuse the textual input into model's representation, we first project the text features from $L\times512$ to $L\times1024$, and then perform cross-attention under this dimension.

For partially-observable experiment (\textbf{P-MO2G}), the target object is guaranteed invisible to the agent initially, but there would be a reference object that is visible to the agent. In the context of navigation, reference object often serves as landmark that the agent can reference to, which assists the agent in inferring the location of the target object. We adopt the same parameters and architecture as the one we have in the \textbf{MO1G} experiment.

\mypar{Baselines.} In detailing the implementation specifics for our baselines within the grid environment, we adhered to a simple yet effective architecture for both Conservative Q-Learning (CQL) and behavioral cloning (BC). This architecture comprises two convolutional layers followed by two fully connected layers. We represent the various objects within the grid as distinct integers to signify the object classes, simplifying the state representation. For our model's hyperparameters, we set the soft update parameter $\tau$ to $0.001$ and the discount factor $\gamma$ to $0.99$, which balances immediate and future rewards. The network architecture features $16$ channels for convolutional processes and a layer size of $40$ for the fully connected layers, ensuring adequate complexity for pattern recognition within the environmental data. The epsilon-greedy strategy for action selection starts with an epsilon value of $1$, annealing to a minimum of $0.01$ over $10,000$ frames to balance exploration with exploitation. The replay buffer is set to a size of $300,000$ to maintain a comprehensive record of past experiences, and the learning rate for the model is configured at $0.001$ to facilitate steady convergence during training.

\subsection{Robotic Arm Block Stacking: Kuka}
\mypar{Ours.} For both stacking and rearrangement experiment. We use a template instruction to describe the final configuration of the cube. An example instruction looks like this:~\textit{Stack green cube on top of red cube. Stack blue cube on top of green cube. Stack yellow cube on top of blue cube.} Our diffusion planner is asked to model the position of the end effector of the robotic arm. We first convert the environment into a 2D grid, which contains the initial configuration of the cube. As a result, the input of our diffusion model contains $3$ channels. The first one is the location of each cube, followed by 2 noisy channels. The first output channel is the trajectories of the end effector, and the second channel is the estimated location of the cube or target placement location.

\mypar{Baselines.} Similarly, for CQL and BC, we have the same parameters as the one we have in the CompILE experiment. We first decompose the stacking task into moving a single cube once. The reward function is designed as the distance to the target position, which initially is the location of the cube to pick, and the location to place the cube (which is often on top of another cube).

\subsection{ALFRED}
\mypar{Ours.} Planning in a complex and realistic environments is extremely challenging. As a result, following~\cite{min2021film, blukis2022persistent}, we first use an object detector to get the semantic map of the agent's field-of-view (FoV) image, together with the depth map from the depth estimation model, we can reconstruct the environment in a 3D point cloud. Each point in the point cloud contains the semantic category of the object. By applying a max-pooling operation across the z-dimension, we derive a 2-dimensional semantic map. In this map, each pixel corresponds to the  object presents at that location. This process aligns with the \textbf{MO1G} setup, where the planning algorithm operates on a 2D plane. Consequently, we employ the same parameters and architecture as in the \textbf{MO1G} experiment. However, to accommodate the full scope of the room, we expand the map dimensions to $48\times48$.

\begin{table}[tp!]
    \centering
    \scalebox{0.92}{
        \begin{tabular}{c|c cc c}
            \toprule
             \multicolumn{1}{c|}{\multirow{2}{*}{\textbf{Models}}}  & \multicolumn{2}{c}{\textbf{Seen}} &  \multicolumn{2}{c}{\textbf{Unseen}}      \\
                                & SR         & GC          & SR              & GC            \\
             \midrule
             EmBERT             & 31.8/23.4  & 39.2/31.3   & 7.5/3.6    & 16.3/10.4      \\
             E.T                & 28.9/20.1  & 36.3/27.8   & 4.7/2.6         & 14.9/8.3      \\
             LACMA              & \textbf{32.4/24.1}  & \textbf{40.5/31.7}   & 9.2/5.8         & 20.1/13.5     \\
             FILM$^{\dagger}$  & 26.7/10.8  & 36.9/14.5   & 22.2/9.6        & 33.9/13.5     \\
             \midrule
             \textbf{Ours}      & 28.7/12.9  & 38.6/16.1   & \textbf{24.7/10.5}       & \textbf{36.3/14.9}     \\
            \bottomrule
        \end{tabular}
        }
    \caption{Results on the ALFRED test set. The path-length weighted results are presented following the slash. ($^{\dagger}$: We test FILM on our machine 5 times and report the highest score.)}
    \label{tab:alfred_test}
\end{table}
\section{ALFRED Test Set Result}
\label{supp:alfred}
Since our method is built upon FILM~\cite{min2021film}, it's essential to showcase the advancements achieved by our planning model over the baseline. However, as reported in the repository of the ALFRED~\cite{shridhar2020alfred} dataset, the simulator of the environment could have machine-dependent behavior, which introduces randomness in the reproducibility of the method. To mitigate this, we conducted five evaluation runs of FILM on our system and report the highest score in~\autoref{tab:alfred_test}. Our model demonstrates a notable improvement in generalization to unseen environments. As indicated in the table, our method achieves a Success Rate (SR) of $24.7\%$ and a Goal Condition (GC) score of $36.3\%$ in unseen scenarios, which is a significant enhancement over the FILM's $22.2\%$ SR and $33.9\%$ GC. This improvement underscores the efficacy of our planning model, particularly in navigating novel environments. 

\section{On the Fly Planning}
The results presented in~\autoref{tab:on_the_fly} reveals the effectiveness of our on-the-fly planning approach on the CompILE dataset under the partially-observable setting. On-the-fly planning refers to the model's iterative process of updating its representation of the environment and adjusting its plan accordingly, continuing until the agent reaches the target or exceed the maximum allowed steps. In the one-shot setting, which model only predicts the path at once, while the Success Rate (SR) is modest at $28.60\%$, the model significantly improves in the two-shot case, achieving an SR of $54.70\%$. Notably, the on-the-fly strategy excels with an impressive SR of $99.9\%$, demonstrating near-perfect performance. Goal Distance (GD) is minimized to 0.002 in the on-the-fly setting, indicating high precision in achieving the specified objectives. Path Precision (Path Prec.), Recall (Path Rec.), and F1 scores reflect this high accuracy and reliability, with the on-the-fly planning outperforming one-shot and two-shot approaches with scores of $93.34\%$, $96.48\%$, and $94.88\%$, respectively. These results underscore the adaptability and efficiency of our on-the-fly planning method when planning under uncertainty.
\definecolor{LightCyan}{rgb}{0.88,1,1}
\newcolumntype{a}{>{\columncolor{LightCyan}}d}

\definecolor{LightGreen}{rgb}{0.9,1,0.9}
\newcolumntype{a}{>{\columncolor{LightGreen}}c}

\begin{table}[tp!]
\centering
\small\addtolength{\tabcolsep}{-1.4pt}
\scalebox{.86}{
    \begin{tabular}{cc|ccccc}
    \toprule
    \multicolumn{1}{c}{\multirow{2}{*}{\textbf{Setting}}} & \multicolumn{1}{c|}{\multirow{2}{*}{\textbf{Models}}} & \multicolumn{5}{c}{\multirow{1}{*}{\textbf{Metrics}}}
    \\
    & &
    SR & GD & Path Prec. & Path Rec. & \multicolumn{1}{c}{\multirow{1}{*}{Path F1}} \\
    \midrule

    \multirow{4}{*}{P-MO2G}
    & \multirow{1}{*}{1-shot}
      & 28.60 & 4.05 & 31.07 & 33.16 & 32.08
    \\
    & \multirow{1}{*}{2-shot}
      & 54.70 & 2.33 & 47.97 & 51.52 & 49.68
    \\
    & \multirow{1}{*}{On-the-fly}
      & \textbf{99.95} & \textbf{0.002} & \textbf{93.34} & \textbf{96.48} & \textbf{94.88}
    \\

    \bottomrule
    
    \end{tabular}
}
\caption{
Analysis of the on-the-fly planning on the CompILE dataset. We evaluate our model under the partially-observable setting to showcase the adaptability of our method when planning under uncertainty.
}
\label{tab:on_the_fly}
\end{table}

\section{More Qualitative Analysis}
\label{supp:qualitative}
\begin{figure*}
    \centering
    \includegraphics[page=5, trim={0 250 200 0}, clip, width=\textwidth]{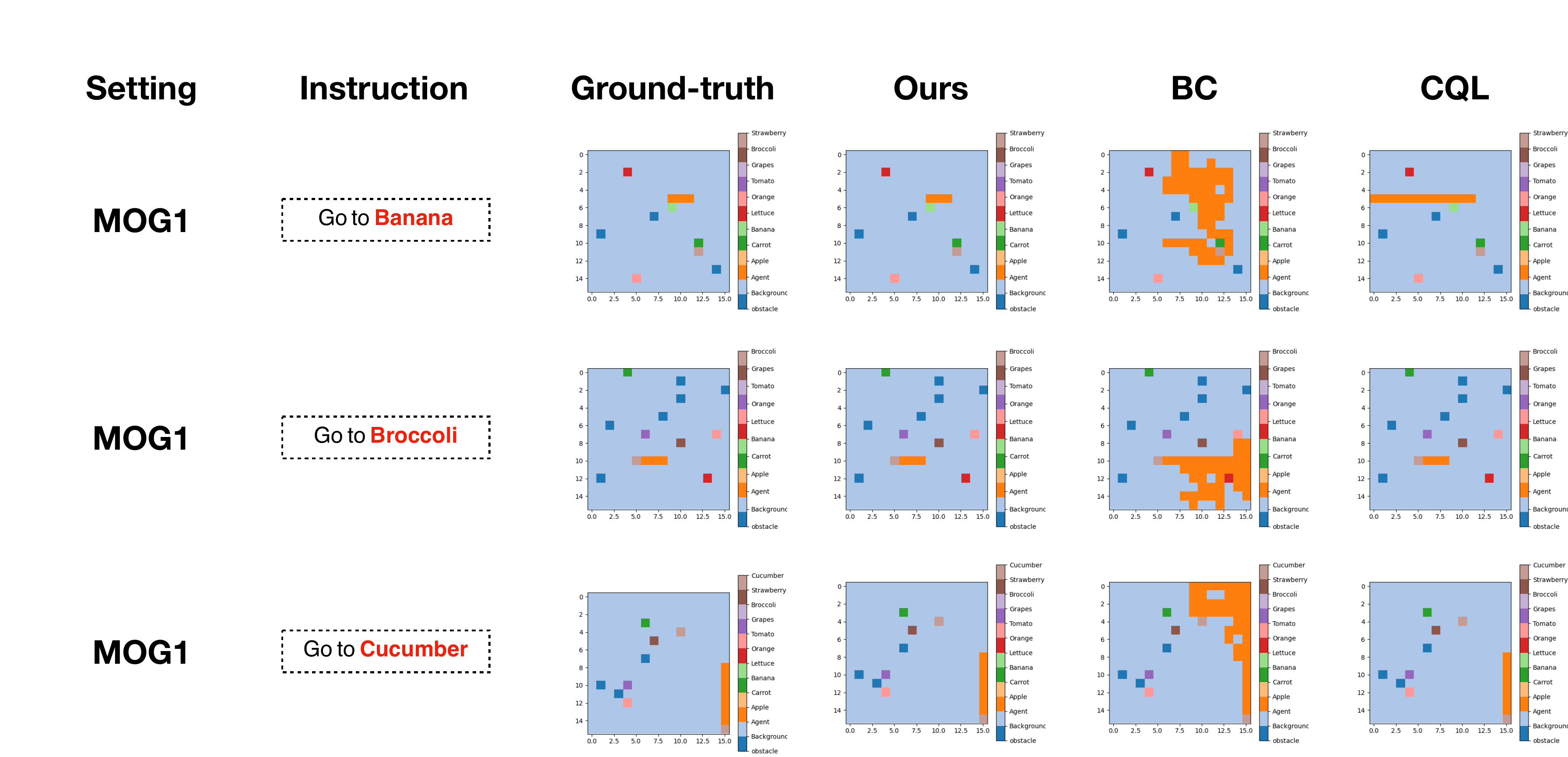}
    \caption{Qualitative Results on CompILE under multi-object and one goal setting. The target objects are marked in bold red.}
    \label{fig:supp1}
\end{figure*}
\begin{figure*}
    \centering
    \includegraphics[page=4, trim={0 250 200 0}, clip, width=\textwidth]{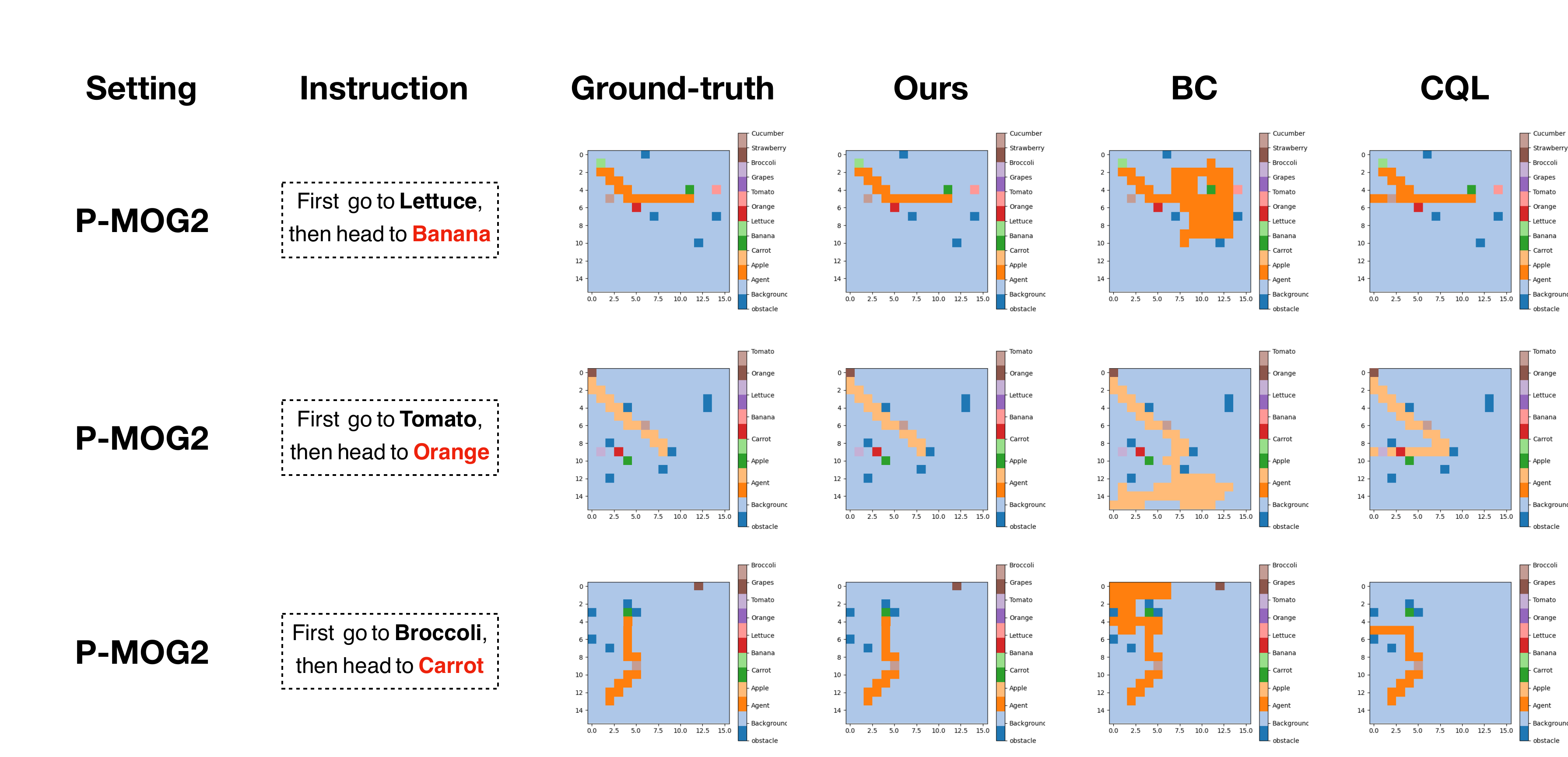}
    \caption{Qualitative Results on CompILE under partially-observable setting. The reference objects are highlighted in bold black and target objects in bold red.}
    \label{fig:supp2}
\end{figure*}
\begin{figure*}
    \centering
    \includegraphics[page=6, trim={0 250 50 0}, clip, width=\textwidth]{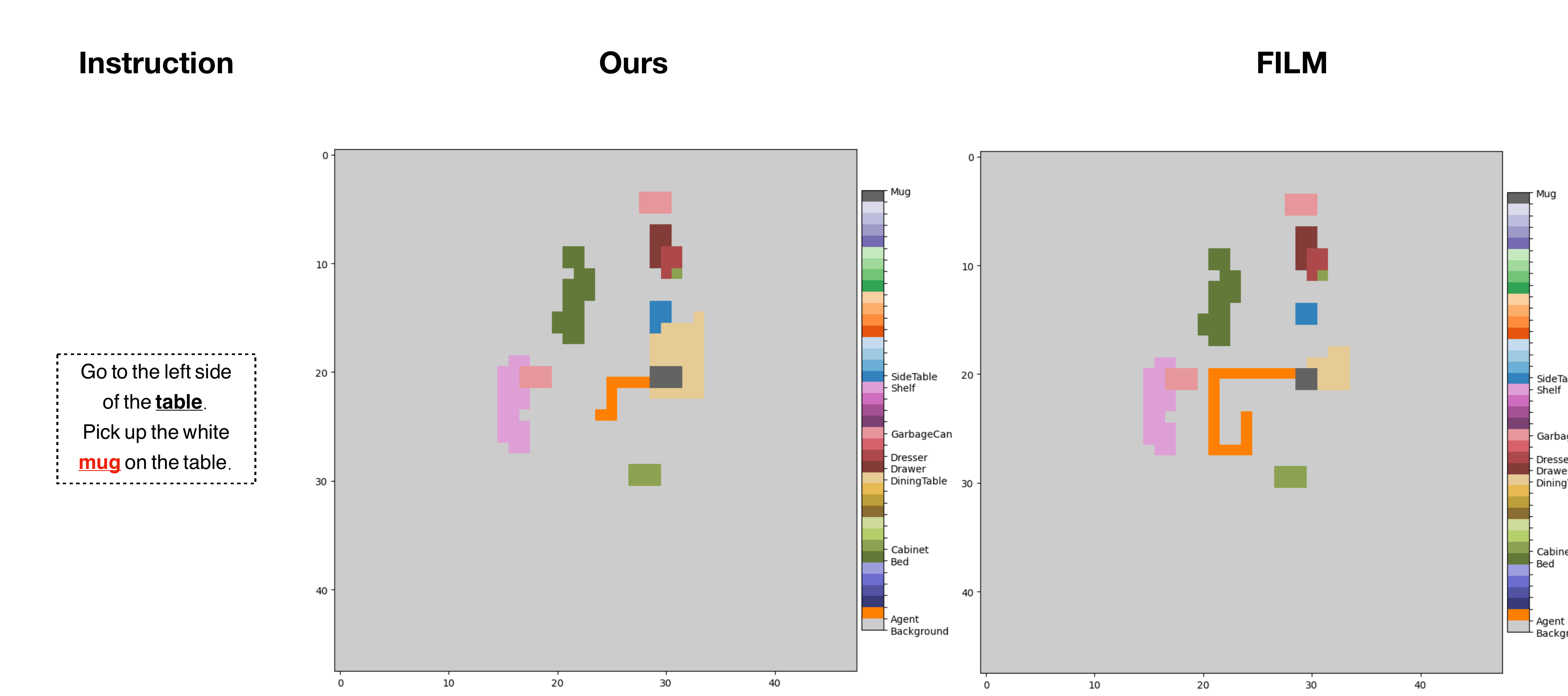}
    \caption{Qualitative Results on ALFRED. The reference objects are highlighted in bold black and target objects in bold red.}
    \label{fig:alfred1}
\end{figure*}
\begin{figure*}
    \centering
    \includegraphics[page=7, trim={0 250 50 0}, clip, width=\textwidth]{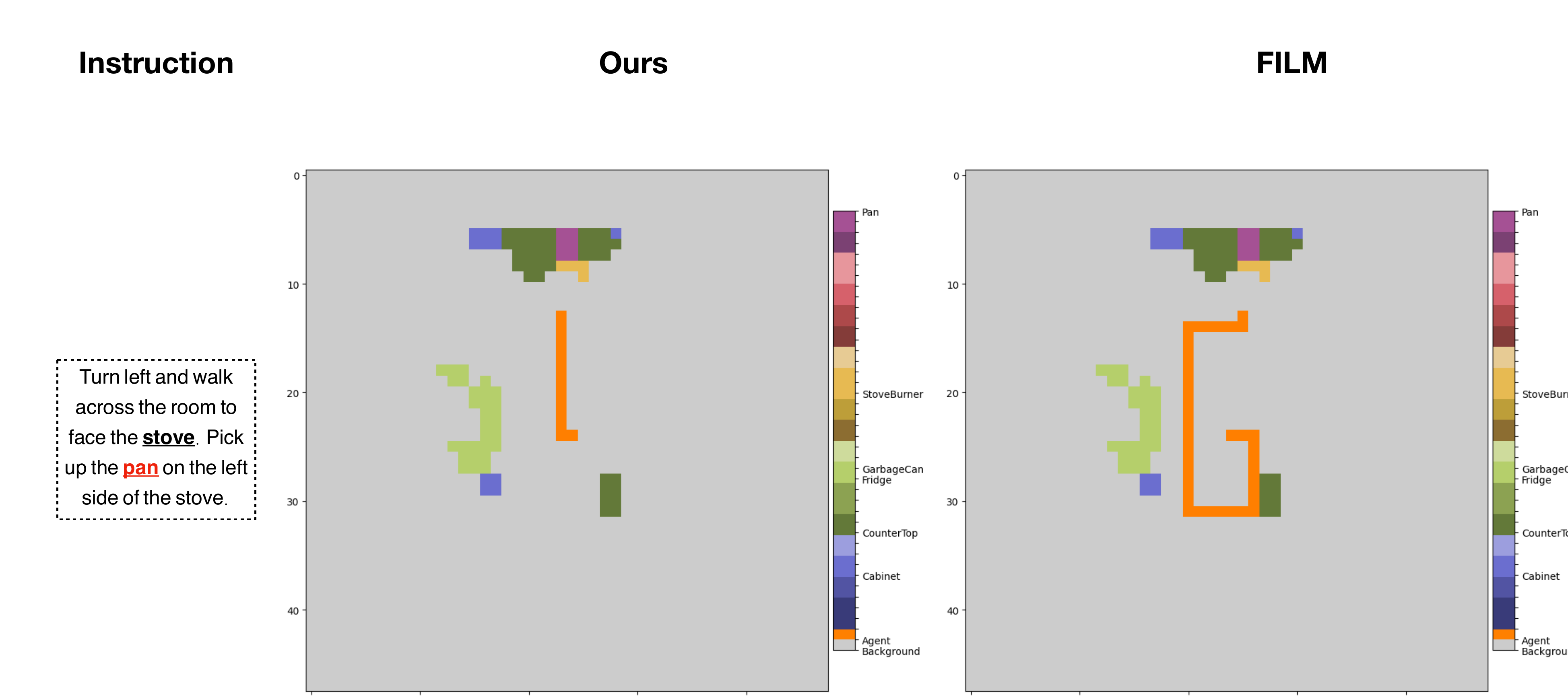}
    \caption{Qualitative Results on ALFRED. The reference objects are highlighted in bold black and target objects in bold red.}
    \label{fig:alfred2}
\end{figure*}

Here we provide more visualization results on the CompILE~\cite{kipf2019compositional} and the ALFRED~\cite{shridhar2020alfred} dataset.

\mypar{CompILE.}
The qualitative results depicted in~\autoref{fig:supp1} offer a comparative analysis of our model's performance against baseline models BC and CQL in the CompILE dataset under two settings: multi-object single goal (\textbf{MOG1}) and partially-observable multi-object goal (\textbf{P-MOG2}). In the MOG1 scenarios, our model demonstrates a high fidelity in replicating the ground-truth paths towards single target objectives such as Banana, Broccoli, and Cucumber, with the targets distinctly indicated in bold red. The baseline models, BC and CQL, display varying degrees of accuracy, with some paths showing indirect routes to the target.

In the more complex partially-observable setting, our model's ability to navigate first to a visible reference object and then to an initially unseen target object is showcased in~\autoref{fig:supp2}. The reference objects serve as waypoints, and our model's trajectories indicate a clear understanding of this two-step process. For example, paths that first go to Lettuce and then head to Banana are more nuanced and exhibit strategic planning that is more aligned with the ground truth compared to the BC and CQL models, which either demonstrate indirect routes or miss the target altogether. The consistent alignment of our model's predicted paths with the ground-truth data across both MOG1 and P-MOG2 settings underscores its robustness and effectiveness in complex task execution.

\mypar{ALFRED.} 
The qualitative evaluations on the ALFRED dataset showcase our model's advanced capability to first identify the reference object and then proceed directly to the target object without the necessity for extensive exploration. For instance, in the task of retrieving the white mug from the table in~\autoref{fig:alfred1}, our model accurately pinpoints the table as the reference object, then moves decisively to the mug. This contrasts with the FILM model, which seems to engage in an initial exploratory phase before locating the mug. Similarly, in~\autoref{fig:alfred2} when tasked with picking up the pan near the stove, our model efficiently recognizes the stove as the reference point and then the pan as the target, following an optimized path that demonstrates a clear understanding of the sequence of actions. The FILM model, however, appears to require a broader search of the environment, indicating a less efficient task execution process. These observations underline our model's proficiency in sequential task execution within complex and partially observable spaces.

\section{Limitations}
\label{supp:limitation}

Our model demonstrates promising results in both navigation and object manipulation tasks; however, it is important to recognize the boundaries within which it operates. We summarize the limitations of our method as follows:

\mypar{Complex and Diverse Task Instructions.} The performance discrepancy observed between the synthetic dataset (CompILE) and the realistic dataset (ALFRED), coupled with the outcomes from the instruction manipulation experiment detailed in Section~\autoref{subsec:analysis}, indicates that the primary challenge for our model is the diversity and complexity of real-world instructions. Addressing this challenge calls for future work to focus on enhancing the model's ability to parse and reason about varied linguistic constructs and the subtleties inherent in real-world instructions.

\mypar{Our planning space is two-dimensional.} Although our model excels at reasoning for certain navigation and manipulation tasks, it operates within a two-dimensional planning space, which restricts its applicability to environments that require three-dimensional spatial reasoning. This limitation presents an opportunity for future research to expand the model's capabilities into the third dimension, allowing for more comprehensive interaction within more complex environments.

\mypar{On-the-fly planning is time-consuming.} While on-the-fly planning did significantly improve our model's performance, the requirement to initiate the diffusion process from scratch for each prediction is a time-consuming procedure. Moving forward, to tackle the issue, we envision two primary avenues for enhancement. On one hand, optimizing the diffusion process for faster convergence or developing methods to resume from intermediate diffusion states could be pivotal. On the other end, we can develop an algorithm to determine whether we need to re-plan or not, which could reduce the computational demands and time costs associated with on-the-fly planning.


\end{document}